\documentclass{INTERSPEECH2023}

\usepackage{tipa}
\usepackage{multirow}
\usepackage{cuted}
\usepackage{capt-of}
\AfterEndEnvironment{strip}{\leavevmode}
\usepackage{pbox}

\interspeechcameraready

\title{Investigating the dynamics of hand and lips in French\\Cued Speech using attention mechanisms and CTC-based decoding}
\name{Sanjana Sankar, Denis Beautemps, Frédéric Elisei, Olivier Perrotin, Thomas Hueber}
\address{
  Univ. Grenoble Alpes, CNRS, Grenoble INP, GIPSA-lab, 38000 Grenoble, France
}
\email{firstname.lastname@gipsa-lab.grenoble-inp.fr}

\begin{document}
\maketitle

\begin{abstract}
Hard of hearing or profoundly deaf people make use of cued speech (CS) as a communication tool to understand spoken language. By delivering cues that are relevant to the phonetic information, CS offers a way to enhance lipreading. In literature, there have been several studies on the dynamics between the hand and the lips in the context of human production. This article proposes a way to investigate how a neural network learns this relation for a single speaker while performing a recognition task using attention mechanisms. Further, an analysis of the learnt dynamics is utilized to establish the relationship between the two modalities and extract automatic segments. For the purpose of this study, a new dataset has been recorded for French CS. Along with the release of this dataset, a benchmark will be reported for word-level recognition, a novelty in the automatic recognition of French CS.

\end{abstract}

\section{Introduction}

 \begin{figure*}[ht!]
 \centering
  \includegraphics[width=0.9\linewidth]{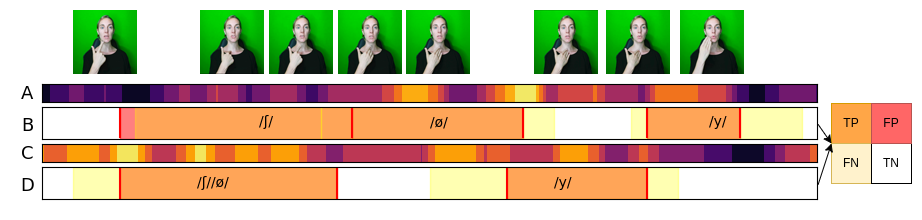}
 \captionof{figure}{Automatic segmentation of hand and lip movements in LfPC from the attention maps of a CTC-based decoder. (A) \& (C) show the attention scores along the DTW-based attention path for lips and hand respectively while (B) \& (D) show the automatic vs. manual segmentation for lips and hand respectively. True positives (TP, in orange) correspond to automatically detected segments that overlap with manually annotated ones. The color scale for (A) \& (C) is as shown in Fig.\ref{fig:dtw}. }
  \label{image_attn}
\end{figure*}

Cued Speech (CS) is a visual communication technique developed by Cornett \cite{cornett} in 1967 to facilitate the interpretation of spoken language for people with hearing impairment. As shown in Figure \ref{image_attn}, it uses a combination of hand shape and hand position with respect to the lips to represent different phonemes being uttered while speaking a language. It helps people who are hard of hearing to differentiate between phonemes which look similar at the lips and are difficult to distinguish while lip-reading, like \textipa{[b]}, \textipa{[p]} and \textipa{[m]}. In this work, the focus is on French CS or Langue fran\c{c}aise Parl\'ee Compl\'et\'ee (LfPC) \cite{cornett_lfpc}, which uses five hand positions to disambiguate vowels and eight hand shapes for consonants.
There have been several attempts to develop systems for automatic CS recognition (ACSR) in recent years to provide speech technologies that aid people with hearing disabilities. In early studies \cite{panikos, aboutabit}, the focus was on isolated vowel and/or consonant recognition and artificial landmarks  are used to facilitate the segmentation and tracking of hand and lips. Later studies have moved on to the more challenging task of \textit{continuous} cued-speech recognition\cite{liu18, liu21, papadimitriou, sankar, liu23}. However, all these studies provide decoding strategies of LfPC at the phonetic level and not at the word level which is a prerequisite for the implementation of a practical solution (e.g. for a dictation application or when combined with a Natural language understanding system). Investigating new decoding strategies of CS at the word level is one motivation of the present study. However, the only available corpus of LfPC is the CSF18 corpus \cite{liu18} which was designed for the purpose of generation (see \cite{guillaume} for its use in the context of Cued-speech synthesis) rather than for recognition. 
Its linguistic material was designed to cover all possible diphones of the French language in a minimum number of sentences. As a consequence, most of the sentences of this corpus present very rare word sequences and unusual syntactic structures, (e.g.``A giant staple may have hit his beautiful speedboat"). These sentences are not suitable for recognition as they are very unlikely for a statistical language model (LM). Also, a new dataset with more realistic linguistic content is needed. To this end, such a dataset has been designed and recorded to be released with this paper. On this dataset, a new model is proposed and evaluated for decoding LfPC at the word level based on (i) automatic hand and lips pose extraction, (ii) RNN-based phonetic decoding (iii) beam search decoding procedure with lexical constraint and n-gram language modeling, and (iv) top-K neural rescoring using a pre-trained GPT2 model.

In addition to this technological objective, this study also aims to analyze such a decoder to better understand the production of CS by humans. In particular, the focus is on the complex dynamic relationships between the hand and lips.   
In fact, there is an inherent asynchrony between the onset of the hand movement w.r.t to the lip movement. In \cite{virginie}, it was established that the hand can precede the lips by a delay of a few milliseconds to several hundred milliseconds. To take into account this asynchrony in ACSRs, \cite{liu18} assumed an average anticipation of the hand and used a heuristic to synchronize the 2 modalities. More recently, \cite{sankar, liu21, liu23} have used bi-directional RNNs to capture useful contextual information for both hand and lips streams before their joint modeling. In this study, it is proposed to use an ACSR system to elucidate the temporal relationships between the hand and lips. To this end, temporal self-attention mechanisms for hand and lip feature streams are integrated within an RNN-based phonetic decoder and the resulting attention maps are analyzed a posteriori. The working hypothesis for this article is the following: in the case of an anticipatory gesture of the hand w.r.t lips, the model should be able to learn to pay attention to the onset of this gesture. Therefore, by detecting any deviation from the diagonal of the attention map, these onsets could be extracted automatically. It should provide a way to visualize hand-lip asynchrony in CS but also to segment hand and lip movements automatically, a time-consuming task and difficult when done manually. Such an automatic segmentation process could thus be useful for the fundamental research on the production and perception of CS \cite{Krause,hage2006effect,machart2020influence,bayard2019}.

In summary, the key contributions of this paper are (i) releasing a new dataset for LfPC adapted to word-based decoding, (ii) providing a first baseline of an automatic decoder of LfPC at the word level and (iii) investigating the use of self-attention mechanisms to provide a fine-grained analysis of hand-lips temporal relationships and a method for segmenting CS data.

\section{Methodology}
\subsection{Model architecture}

\begin{figure*}[t!]
  \centering
  \includegraphics[width=0.9\linewidth]{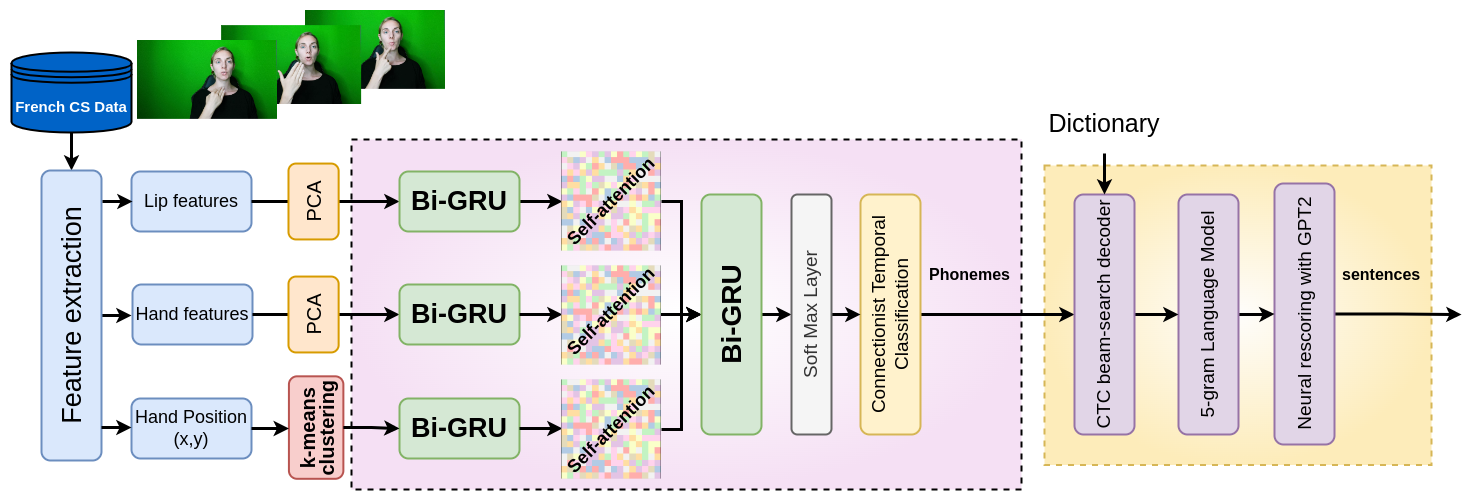}
  \caption{Model architecture}
  \label{fig:arch}
\end{figure*}

The model architecture for the ACSR task is presented in Fig.\ref{fig:arch}. Similarly to \cite{sankar}, it has 3 streams corresponding  to lips, hand shape and hand position respectively. 
For each stream individually, a first Bidirectional Gated Recurrent Unit (Bi-GRU) captures contextual information from the sequence of visual features. However, to interpret how the model exploits this contextual information, a temporal self-attention mechanism is added after each (Bi-GRU). This mechanism is based on the scaled-dot product attention defined in \cite{vaswani} as: 
\begin{equation}
    Attention(Q, K, V) = softmax(\frac{QK^T}{\sqrt{d_k}})V = W_{att} V
\end{equation}
where K, Q, and V (key, query, and value) are embedded representations learned from the input sequence (in our case, the output of the Bi-GRU). Here, $W_{att}$ is the attention score at each time step for all the frames of a given sequence.
The outputs of each self-attention mechanism are then concatenated together, passed to a Bi-GRU (processing the 3 streams of information jointly) and finally to the output \textit{softmax} layer. The decoder is trained only with phonetic sequences as a target using the CTC loss \cite{graves} and importantly, without any temporal segmentation at the phonetic level (the dimension of the final softmax layer is $k+1$ where $k$ is the number of phonetic classes, one output encoding the CTC \textit{blank} token).

\subsubsection{Word-level decoding}
Word-level decoding is achieved using a beam search decoder. During beam search, phonetic hypotheses are iteratively expanded with the next possible phonemes. At each time step, only the hypotheses with the highest scores are retained. 
A pronunciation dictionary associating words with phonetic transcription is used to constrain the decoding (only words from the dictionary can be generated). An N-gram language model provides a likelihood score for each sequence of hypotheses during the beam-search decoding. An auto-regressive neural model based on Transformer (GPT) is finally used a posteriori to re-score the top-K sequences produced by the beam search decoder. This re-scoring is done by computing its perplexity  w.r.t each candidate sequence $X^{(k)} = (x_0,\ldots,x_N)$ (classically tokenized), defined as:
\begin{equation}
\operatorname{PPL}(X^{(k)})=\exp {-\frac{1}{t} \sum_i^t \log p_\theta(x^{(k)}_i \mid x^{(k)}_{<i})}
\end{equation}
where $\theta$ is the parameter set of the LM. The candidate sequence for which the perplexity is minimal is finally selected.

\subsection{Attention-based temporal segmentation}
\label{sec_dtw}
In accordance with the working hypothesis, the model should pay attention to the onset of hand and/or lips movements, providing a way to segment those movements. Therefore, for a given test sentence, the attention maps for hand and lip streams are extracted. From these attention maps, an optimal path - hereon referred to as the attention path -  along which the cumulative attention is maximum is extracted using the Dynamic Time Wrapping (DTW) algorithm (the attention score between each pair of frames of the sequence is used as the local distance). A Sakoe-Chiba band of 30 frames is used to constrain the attention path to be not too far from the diagonal of the attention map. Frames at which the attention map deviates from the diagonal are expected to match the onset of hand/lip movements. This simple heuristic also avoids setting an arbitrary threshold on the value of the attention scores. When there are several consecutive frames detected as candidates for the movement onset, the middle frame is considered. We then constrain the number of segments to be not more than the number of predicted phonemes and assign the boundaries to the predicted phonemes. Finally, the respective position of the onsets of hand and lip movements for each phoneme reveals the asynchrony between these two modalities of CS. 

For evaluating the accuracy of the proposed  segmentation technique, ``temporal intersection over union" $tIoU$ is calculated as described in \cite{visuallygrounded} and defined as: 
\begin{equation}
tIoU = \frac{1}{|S|} \Sigma_{s \in S} \frac{l_{ts} \cap l_{ps}}{l_{ts} \cup l_{ps}},
\label{eq:iou}
\end{equation}
where $S$ is the set of all the segments from manual annotation (ground truth), $l_{ts}$ is the true annotation segment $s$ is assigned to, and the output of the intersection (or union) operation is the number of frames of intersection (or union) between the segment spanned by $l_{ps}$ which is the predicted segment.\\

\section{Experiments}
\subsection{CSF2022 Dataset}
The new dataset consists of 1\,087 sentences recorded by a professional French cuer with typical hearing. These sentences were selected from the SIWIS database \cite{honnet2017siwis} and are originally from various sources, such as French parliament debates, French novels and semantically unpredictable sentences (the latter not being  used at test time for word-based decoding). The experimental setup also includes LED lights (placed next to the cuer's face) and a green background. The cuer was also requested to utter the sentences while cueing. The sentences were prompted on the screen using a dedicated web-based application. In the mid-term perspective of recording data 'in the wild', i.e. professional CS interpreters recording data in various environments and devices (e.g. at home with their smartphone), the videos were recorded  using a simple webcam (1920x1080 pixels). The targeted framerate was set to 60 fps. Audio was recorded at 44.1 kHz (PCM, 16bits) but is not used in this study. In total this new dataset contains 97mn of CS data (i.e. 3 times more than the CSF18 corpus). 
 The phonetic transcription  of the recorded sentences was obtained automatically using the LIA-Phon phonetizer \cite{liaphon} and verified/corrected manually (36 phonetic categories are used to describe French). 
To evaluate the accuracy of the attention-based segmentation, a set of 20 sentences was manually annotated with the ELAN software \cite{elan} to provide the true segmentation boundaries for lips, hand shape and hand position based on the video. 
The dataset (including raw data, hand/lips pose estimation,  manual annotations, and dedicated train and test sets) will be made available publicly in association with this paper. 
 
\subsection{Visual Features}
\label{sec_feats}
As mentioned above, the videos are recorded with a target framerate set to 60fps via a web application that facilitates the recording of data in the wild. However, due to the use of a webcam (and probably also due to the encoding codec), the frame rate is found to be variable when extracting the images (i.e. 60Hz with a few missing frames). 
Each extracted image is then processed by the Mediapipe software \cite{mediapipe} which provides 21 x-/y-coordinates for hands and 42 x-/y-coordinates for the lips. The resulting sequence of visual feature is re-interpolated to match the rate of 60 fps. 
The dimension of the visual feature vectors is then reduced to 20 for each stream using PCA (explaining more than 99\% of the variance). For the hand position features, the data is clustered using the k-means  algorithm and the cluster labels (one-hot encoded) are used as the feature for the third stream. 8 distinct clusters are chosen to account for the 5 hand positions and a few transition states.

\subsection{Implementation details}
Each Bi-GRU of the ACSR model has 256 units. The embedding size of the weight matrix used to obtain the Q, K and V matrices in the (single-head) self-attention layer is set to 256. 
The Adam optimizer is initialized with a learning rate of 0.001 and reduced on a plateau. The model is trained for 120 epochs with a batch size of 16.

For the word-level decoding, a pronunciation dictionary is created with all the words in the new CSF2022 dataset to which a list of the 1000 most frequent words used in French is added. This list was extracted from \textit{Spacy.io} online tool, resulting in a 3.2k words lexicon. For the beam search decoding, the pre-trained 5-gram LM provided in the Voxpopuli corpus \cite{wang-etal-2021-voxpopuli} is used. Evaluations are conducted on a test set of 108 sentences selected randomly (avoiding semantically unpredictable sentences). To the best of our knowledge, these sentences are not in the training set of this LM (which is rather based on European Parliament debates). The CTC beam search decoder available in PyTorch (version 1.13.1) is used in this study. The hyperparameters \textit{beam\_width},  \textit{word\_score} and \textit{lm\_weight} are set respectively to 1000, 0 and 0.2, after being optimized on a subset of the training set of 30 sentences. For the rescoring procedure, the top-30 sequences of words from the beam search decoder are considered and the pre-trained GPT2 model for French proposed by \cite{simoulin:hal-03265900} and available on \textit{HuggingFace} is used.

\section{Results and Discussion}

\subsection{Recognition performance}

\begin{table*}[t]
    \caption{Performances of the proposed system for automatic decoding of French Cued-Speech,  both at the phonetic and word levels, in terms of correctness (Corr) and accuracy (Acc, which takes into account the insertion errors). Here, $Acc.=100(1 - WER)\%$. The Wilson confidence interval $\Delta_{95\%}$ is also reported, assuming a binomial distribution of the errors.}
    \label{tab:decoding_performance}
\centering
    \begin{tabular}{l c c c}
      \hline
      \textbf{Configuration} & \textbf{Level} & \textbf{Corr $ \pm \Delta_{95\%}$} & \textbf{Acc $ \pm \Delta_{95\%}$} \\ 
      \midrule
      Baseline phonetic decoder \cite{sankar} & Phone &  81.3 $\pm$ 1.6   & 77.5 $\pm$ 1.8  \\ \hline
       Phonetic decoder (baseline + self-attention)  & Phone & 81.2 $\pm$ 1.6 & \textbf{76.6} $\pm$ 1.8  \\ \hline
      \multirow{2}{*}{Phonetic decoder + Lexicon} & Phone & 77.0 $\pm$ 1.6  & 74.1 $\pm$ 1.8   \\
      & Word & 41.8 $\pm$ 3.9 & 27.7 $\pm$ 3.9 \\ \hline
      \multirow{2}{*}{Phonetic decoder + Lexicon + 5-gram LM} & Phone & 79.4 $\pm$ 1.6  & 76.8 $\pm$ 1.8 \\
      & Word & 54.5 $\pm$ 3.9 & 52.9 $\pm$ 3.9 \\ \hline
      \multirow{2}{*}{Phonetic decoder + Lexicon + 5-gram LM + GPT-based rescoring} & Phone & 79.2 $\pm$ 1.6  & 76.5 $\pm$ 1.8  \\
      & Word & \textbf{61.9} $\pm$ 3.9 & \textbf{58.5} $\pm$ 3.9  \\ \hline
      \multirow{2}{*}{Phonetic decoder + Lexicon + Dict + 5-gram LM + 30-best} & Phone & 82.8 $\pm$ 1.5  & 80.4 $\pm$ 1.6  \\
      & Word & 71.4 $\pm$ 3.5 & 70.7 $\pm$ 3.5 \\ 
      \bottomrule
    \end{tabular}
    
\end{table*}

Since the proposed system is partially based on the phonetic decoder described in \cite{sankar}, the latter is used as a baseline and its performance on the new corpus is evaluated. 
The results are presented in Table \ref{tab:decoding_performance}. 
First, thanks to the temporal attention mechanisms, the phonetic decoder is more easily interpretable while also maintaining performance as the baseline(Acc=77.5\% vs. 76.6\%). 
Several ablations studies are conducted to assess the impact of the different modules involved in the decoding at the word level. 
It can also be observed that constraining the vocabulary slightly degrades the performance at the phonetic level (Acc=74.1\% vs. 76.6\%). This is mainly due to a higher number of deletion errors 
and can be explained by the lack of information about the ``liaison", which occurs quite often in the French language. This issue is alleviated when using the 5-gram LM (Acc=76.8\%).

Best performances at the word level are obtained by combining the 5-gram model and the GPT-based rescoring, (Acc=58.5\%). It can be noted that the 5-gram model improves the accuracy by 25.2\% compared to the lexicon-only configuration (Acc=52.9\% vs. 27.7\%) and that the GPT-rescoring resulted in an additional 5.6\% improvement (Acc=58.5\% vs. 52.9\%). The analysis of the decoded sequences reveals that an important number of errors are due to deletion or insertion of short words (e.g.``vous \textbf{n'}avez donc rien inventé" instead of ``vous avez donc rien inventé \textbf{et}", or syntax errors (e.g. ``à leur famille" instead of ``à leur\textbf{s} famille\textbf{s}") which are unfortunately not always corrected by GPT-2. To confirm these interpretations, the performance obtained when considering the 30-best hypotheses from the beam search decoding is reported while keeping only the one for which the accuracy at the word level is maximum. The performance is much higher (Acc=70.7\% at the word level) but obviously, this configuration cannot be used in practice since the ground truth is not known at decoding time. However, it provides an estimate of the performance that could be achieved if ambiguities at the semantic level could be eliminated (e.g. by exploiting a larger linguistic context).

\subsection{Attention-based segmentation}
\begin{figure}[h]
  \centering
  \includegraphics[width=0.9\linewidth]{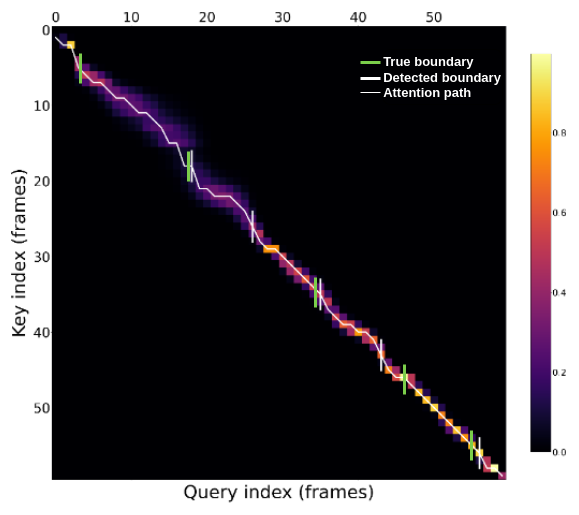}
  \caption{Attention map for hand shape with corresponding DTW attention path and detected movement onsets (for one sentence from the test set)}
  \label{fig:dtw}
\end{figure}

Fig.\ref{fig:dtw} shows the onset of hand shapes as detected from the attention map along the attention path and from manual annotation.
It can be seen from the graphs that the detected segmentation onsets are close to the manually annotated onset. On average, the automatically detected segmentation onset falls within 5 frames of the ground truth (i.e $\pm$ 40ms). 
Similar results are observed for the lips, but surprisingly, the attention map for the hand position is much less interpretable. This needs further study and the focus here is on lips and hand shape.

In Fig.\ref{image_attn}, an example of automatic segmentation of both lips and hand movements is presented for one of the manually annotated sentences.  
While not perfect, it is clear that the proposed method segments the visual data in a consistent manner w.r.t to the manual annotation. This tendency is confirmed by the relatively high $tIoU$ (averaged over all the manually annotated sentences, see Eq.\ref{eq:iou}) which was found to be 60.3\% for hand and 69.6\% for the lips. From the analysis of the attention maps and from the automatic segmentation results, it can be observed that the temporal relationships between the hand and lips in CS are complex.  The hand gesture may precede the lip gesture by several hundred milliseconds. However, the two articulators are sometimes required to synchronize. In such cases, the two attention paths overlap and are close to the diagonal. 

\section{Conclusions}
Although the decoding performances at the word level are still quite far from those of an automatic acoustic speech recognition system, the proposed ACSR system is a first step toward a system that can be used in practice. In addition, it is shown that the attention maps of an ACSR can be used to automatically detect (with a minimum of post-processing) the onset of hand and lips movement in Cued-speech, providing potential useful information on its underlying organisation. 
In our future work, this technique will be applied to a multi-cuer database, in order to study potential speaker-specific strategies in hand-lips coordination in cued speech.  
\section{Acknowledgements}

\ifinterspeechfinal
     This work, as part of the Comm4CHILD project, has received funding from the European Union’s Horizon 2020 research and innovation program under the Marie Sklodowska-Curie Grant Agreement No 860755. The authors would like to thank the CS cuer recorded for this study and the Ives company for providing the web-based recording tool. 
\else
     The authors would like to thank the CS cuer recorded for this study.
\fi

\bibliographystyle{IEEEtran}
\bibliography{mybib}

\end{document}